\documentclass[preprint,12pt,authoryear]{elsarticle}

\graphicspath{ {./figures/} }
\usepackage{hyperref}
\usepackage{float}
\usepackage{verbatim} %comments
\usepackage{apalike}
\usepackage{multirow}
\usepackage{amsmath}
\usepackage{amsthm}
\usepackage{amssymb}
\usepackage{caption}
\usepackage{subcaption}
\usepackage{xcolor}

\usepackage{inconsolata}
\usepackage{graphicx}
\usepackage{multicol}
\usepackage{multirow}
\usepackage{xcolor}
\usepackage{comment}
\usepackage{algorithm}
\usepackage{algpseudocode}
\usepackage{tcolorbox}

\journal{Knowledge-Based Systems}

\begin{document}

\begin{frontmatter}



\title{The Lay Person’s Guide to Biomedicine: Orchestrating Large
Language Models\tnoteref{label1}}


\author[label1]{Zheheng Luo}
\ead{zheheng.luo@manchester.com}
\author[label1]{Qianqian Xie}
\ead{qianqian.xie@manchester.com}
\author[label1]{Sophia Ananiadou}
\ead{sophia.ananiadou@manchester.com}
\affiliation{[label1]organization={The University of Manchester},%Department and Organization
            addressline={Oxford Road}, 
            city={Manchester},
            postcode={M13 9PL}, 
            state={},
            country={United Kingdom}}

%\affiliation{[label3]organization={Yale University},%Department and Organization
%            country={United States}}

\begin{abstract}
Automated lay summarisation (LS) aims to simplify complex technical documents into a more accessible format to non-experts.
Existing approaches using pre-trained language models, possibly augmented with external background knowledge, tend to struggle with effective simplification and explanation. Moreover, automated methods that can effectively assess the `layness' of generated summaries are lacking. Recently, large language models (LLMs) have demonstrated a remarkable capacity for text simplification, background information generation, and text evaluation.  This has motivated our systematic exploration into using LLMs to generate and evaluate lay summaries of biomedical articles.
We propose a novel \textit{Explain-then-Summarise} LS framework, which leverages LLMs to generate high-quality background knowledge to improve supervised LS.
We also evaluate the performance of LLMs for zero-shot LS and propose two novel LLM-based LS evaluation metrics, which assess layness from multiple perspectives.
Finally, we conduct a human assessment of generated lay summaries.
Our experiments reveal that LLM-generated background information can support improved supervised LS. Furthermore, our novel zero-shot LS evaluation metric demonstrates a high degree of alignment with human preferences. We conclude that LLMs have an important part to play in improving both the performance and evaluation of LS methods.

\end{abstract}


\begin{highlights}

\item Large language models show promise in simplification and evaluation while existing methods struggle.
\item "Explain-then-Summarise" framework proposed to enhance LS with LLM-generated background knowledge.
\item Human assessment confirms LLMs improve supervised LS and evaluation metrics.
\end{highlights}


\end{frontmatter}


\section{Introduction}
To make biomedical knowledge more accessible to a non-expert (or \textit{lay}) audience   \citep{10.7554/eLife.27725,Crossley2014WhatsSS}, various automated lay summarisation (LS) methods have been proposed
   \citep{Guo2021,Devaraj2021}. These methods aim to distil the knowledge contained within technical documents into plain language summaries that can be easily digested by the general public.
A popular approach to LS involves fine-tuning pre-trained language models (PLMs) using lay summarisation datasets   \citep{Guo2021}; knowledge retrieved from external databases can support these methods in generating comprehensive background information   \citep{Guo2022}.
Generating an `ideal' lay summary involves two important steps, i.e., \textbf{(i) linguistic simplification} - removing technical jargon, splitting/simplifying complex syntactic structures, and inserting cohesive devices   \citep{Crossley2012TextSA}; and \textbf{(ii) background explanation generation} - generating additional background knowledge not present in the original document to enhance lay readers' comprehension   \citep{Crossley2014WhatsSS}.
Although previously proposed models have begun to lay the foundations for effective LS, three main challenges remain unresolved.  

\begin{figure*}[!tb]
    \centering
    \includegraphics[scale=0.5]{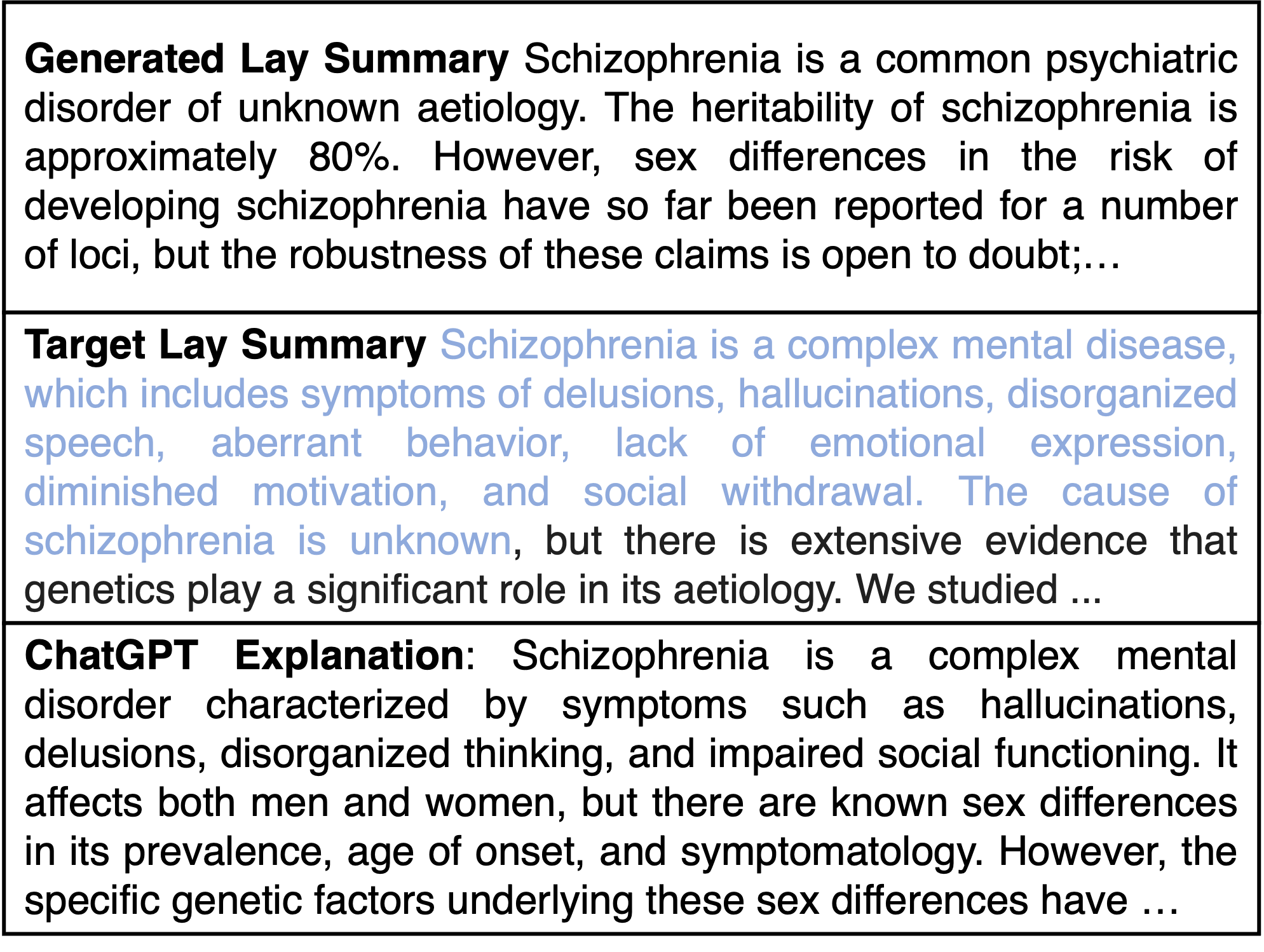}
    \caption{Abstract and human-authored lay summary from  \citet{Shifman2008GenomeWideAI}. The blue text in the lay summary denotes background information absent from both the abstract and the full paper. The bottom section shows the explanation of the abstract generated by ChatGPT, which is highly aligned with the background information in the lay summary.}
    \label{fig: knowledge gap}
\end{figure*}

Firstly, existing LS models   \citep{Guo2021,Goldsack2022,Luo2022ReadabilityCB} often struggle to carry out effective \textbf{linguistic simplification}; their ability to do so is significantly impacted by the degree to which the target, human-authored lay summaries in the training data have been simplified. Overly complex target summaries can hinder the models' ability to perform sufficient simplification   \citep{Luo2022ReadabilityCB}, while summaries that are too simple may cause models to over-correct, resulting in grammatical and typographic errors   \citep{Devaraj2021}.
Secondly, effective \textbf{background explanation generation} of technical terms is challenging, since such explanations are absent from training documents (see Fig. \ref{fig: knowledge gap}).
%shows an example concerning “schizophrenia”, where the generated lay summary from  \citet{Luo2022ReadabilityCB} failed to contain the background explanation for this technical term since the source training document did not provide any background information about "schizophrenia". 
As a solution,   \citet{Guo2022} augmented technical training documents with background information retrieved from external knowledge bases. Despite positive results, the labour-intensive nature of building an external knowledge base and the difficulties in retrieving high-quality, relevant information limit the utility of this approach.

Finally, there is a lack of automated methods for \textbf{evaluating lay summaries} that account for multiple ``layness'' facets, such as linguistic simplification and background explanation. Traditional readability metrics like the Coleman-Liau Index (CLI)   \citep{Coleman1975} and recently proposed PLM-based metrics   \citep{Devaraj2021, Luo2022ReadabilityCB} only consider shallow textual features.

Recently, Large Language Models (LLMs)   \citep{brown2020language} have demonstrated promising performance in various natural language processing (NLP) tasks   \citep{Stiennon2020, Kojima2022} that are highly relevant to solving the issues with LS outlined above. Firstly, LLMs can perform text simplification with human-level accuracy   \citep{lyu2023translating, Shaib2023SummarizingSA}. Secondly, they can generate high-quality background information   \citep{AlKhamissi2022ARO,yu2022generate}, as exemplified in the bottom part of Fig.\ref{fig: knowledge gap}.
%; such information could potentially support more effective LS. 
Finally, they can act as effective evaluators of automatically generated text   \citep{Luo2023ChatGPTAA, Kocmi2023}, given appropriate hand-crafted instructions. 
%Accordingly, LLMs demonstrate significant potential to solve the issues of existing LS methods that were introduced above.  T
To further investigate the potential of LLMs in generating and evaluating lay summaries, this study aims to answer the following research questions: 
\textbf{Q1:}  \textit{Can LLMs generate useful background knowledge that can improve existing LS methods?}
\textbf{Q2:}  \textit{How well can LLMs perform LS in a zero-shot setting?} 
\textbf{Q3:}  \textit{Can LLMs act as effective LS evaluators?}
%Our main contributions, which result from responding to these questions, are depicted in Fig \ref{fig:pipeline}, and can be summarised as follows: 

To respond to these questions, we comprehensively explore LLMs for both the generation and evaluation of lay summaries.
To answer Q1, we propose a novel supervised fined-tuned framework for LS, \textit{Explain-then-Summarise (\textbf{ExpSum})}. This framework augments technical articles with LLM-generated background knowledge about technical terms prior to fine-tuning, as means to support more effective LS. In contrast to existing methods that retrieve knowledge from external databases   \citep{Guo2022}, our method requires neither an external knowledge base nor the development of task-specific training data, since background knowledge is obtained using zero-shot prompting of LLMs. Furthermore, our experiments show that LLM-generated background knowledge is generally of higher quality than knowledge obtained using retrieval-based methods.
%due to the remarkable capability of LLMs to generate text and to exploit the world knowledge that they possess. 

To respond to Q2 and Q3, we explore the performance of LLMs in carrying out zero-shot LS and propose two LLM-based evaluation methods, using a prompt that instructs LLMs to consider multiple facets important for effective lay summary generation and evaluation. 
%To adres%we explore the suitability of LLMs to evaluate the `layness' of generated summaries, 
%we propose two LLM-based evaluation methods that take into account a wider range of features that are important for evaluating lay summaries, including lexical simplification, avoidance of technical details, and clear explanations of complex terms.
Finally, we propose the first detailed human evaluation protocol for LS evaluation, which we use as the basis for the human assessment of lay summaries generated by different methods.
Our experimental results on two benchmark LS datasets and human evaluation results reveal that: 

1) ExpSum significantly outperforms baseline LS methods, thanks to the high-quality background knowledge generated by LLMs. 
%compared to retrieval methods, showing LLMs an efficient explanation generator for improving LS.

2) LLMs exhibit significant potential for zero-shot LS, with a performance level that is comparable to that of supervised methods.

3) The performance of supervised methods in comparison to zero-shot LLM prompting depends on the characteristics of target lay summaries in the training dataset. Supervised methods outperform zero-shot prompting when target summaries are highly simplified, but the reverse is true when simplification is less extreme. 

4) The performance of our novel ChatGPT-based LS evaluation metric matches or even surpasses the effectiveness of traditional and other LLMs-based metrics, emphasising the potential of ChatGPT for effective LS evaluation.

%4) Traditional metrics are insufficiently in alignment with human preferences on assessing lay summaries, highlighting the necessity of efficient metrics tailored for LS.

\begin{figure*}
    \centering
    \includegraphics[width=1.0\textwidth]{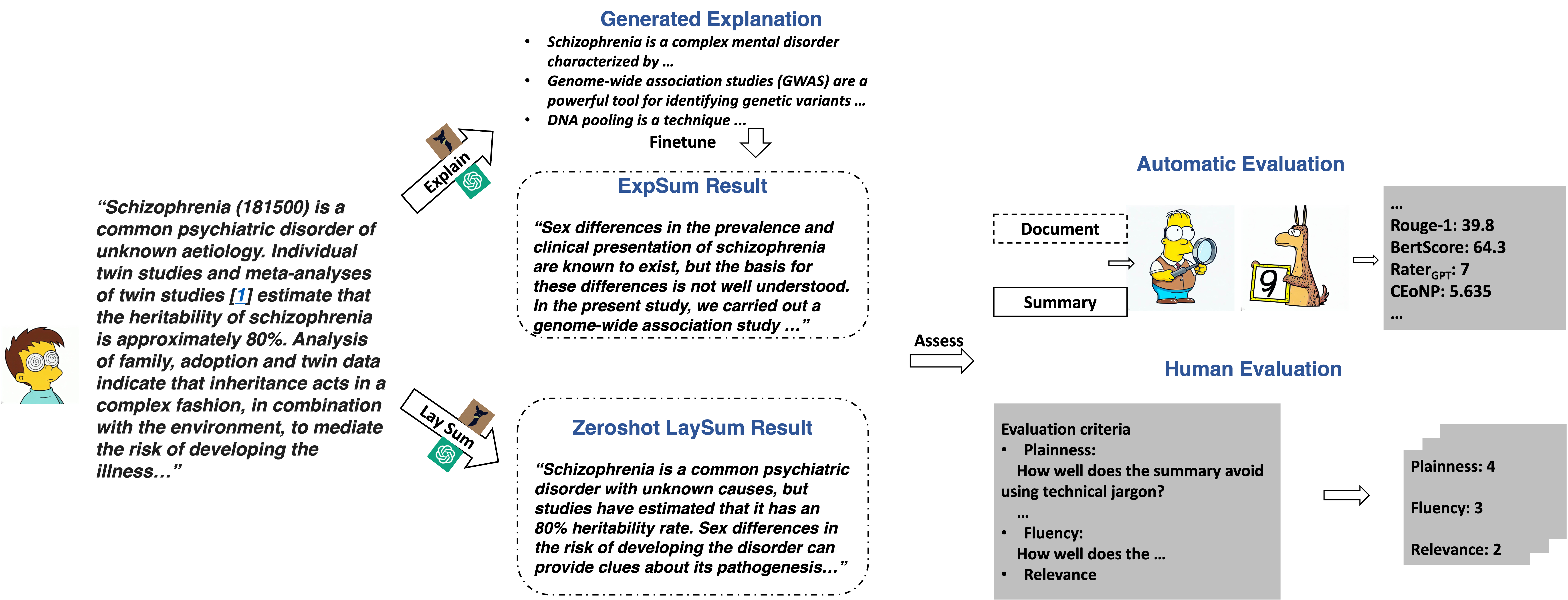}
    \caption{LLMs for LS: in background explanation, summary generation and evaluation.}
    \label{fig:pipeline}
\end{figure*}

Our main contributions are graphically depicted in Fig \ref{fig:pipeline} and can be summarised as follows:
1) We propose a novel Explain-then-Summarise (ExpSum) framework for LS, leveraging LLMs for background explanation, 2) We examine the ability of LLMs for zero-shot LS, 3) We propose two LLM-based "layness" evaluation metrics, 3) We propose the first evaluation protocol for human assessment of LS, 4) We provide insights into the advantages and disadvantages of LLMs for LS and existing evaluation metrics.
%Notably, enriched by background information from LLM, a finetuned model can outperform their convention counterpartss and retrieve-augmented methods.
%\item We further explored zero-shot plain language summary generation using LLMs and compare them against output from finetuned models. We designed a plainness evaluation framework and conduct human assessment of plainness. We found ...

\section{Related Work}
\paragraph{Lay summaristion}
There has recently been a surge of interest in LS, initially driven by the LaySum track in the CL-SciSumm 2020 shared task series   \citep{Chandrasekaran2020OverviewAI}.
%, where most participating teams fine-tuned generative language models such as PEGASUS   \citep{zhang2020pegasus} and BART   \citep{lewis2020bart} on the small corpus of 572 training developed for the task. 
%   \citet{Devaraj2021} framed LS as the paragraph-level simplification of systematic reviews and applied \textit{unlikelihood training}   \citep{Welleck2019NeuralTG} to generate summaries with lower readability scores.
Subsequently, \citet{Guo2021} collected 7K systematic reviews with their respective lay summaries and used these to evaluate the performance of SOTA summarisation methods, including BERT  \citep{kenton2019bert} and BART  \citep{lewis2020bart}.
  \citet{Luo2022ReadabilityCB} and   \citet{Goldsack2022} gathered a larger dataset of PLOS biomedical articles, accompanied by their technical abstracts and lay summaries, and used them to benchmark mainstream PLM methods, including BART and LongFormer  \citep{beltagy2020longformer}.
  \citet{Guo2022} collected 63K scientific abstracts and their corresponding lay summaries from 12 different journals and proposed a retrieval-augmented method (RALL) to acquire external information that bridges the knowledge gap between abstracts and lay summaries.

\paragraph{LLMs for Context Generation}
Recent research has demonstrated that substantial knowledge can be acquired from LLMs, particularly for tasks requiring contextual knowledge, e.g., commonsense reasoning tasks  \citep{Liu2021GeneratedKP}. 
  \citet{Yu2022GenerateRT} prompted LLMs to generate contextual documents based on a given question and combined them to produce a final answer. Their experiments on open-domain QA, fact-checking, and
dialogue systems demonstrate that LLM outputs significantly outperform previous retrieval methods in knowledge-intensive tasks.

\paragraph{LLMs for Text Evaluation}
Recently proposed LLM-based metrics can assess specific aspects of generated texts, such as factuality, consistency and fluency. GPTScore   \cite{Fu2023GPTScoreEA} uses the likelihood that a sequence will be generated by a language model, conditioned on a prefix that specifies the tested aspect, as a means of evaluating the aspect. Further studies have shown that LLMs can achieve human-level performance in the instruction-guided rating of machine translation quality   \cite{Kocmi2023} and in evaluating factual consistency   \cite{Luo2023ChatGPTAA}.

\section{LLMs for Lay Summarisation and Evaluation}
In this section, we firstly describe our \textit{Explain-then-Summarise (ExpSum)} framework, which aims to improve LS by prompting LLMs to provide background knowledge. 
%and answers the research question \textbf{Q1}.
We subsequently describe the use of  LLMs for zero-shot LS and finally explore how they can be used to evaluate lay summaries. 
%to reply the research question \textbf{Q2} and \textbf{Q3}.

\subsection{Explain-then-Summarise (ExpSum)}
\noindent\textbf{Explain} Given an LS dataset consisting  of a document set $D=\{d_1,d_2,\cdots,d_k\}$, a technical abstract set $S^{abs}=\{s^{abs}_1,s^{abs}_2,\cdots,s^{abs}_k\}$, and a lay summary set $S^{pls}=\{s^{pls}_1,s^{pls}_2,\cdots,s^{pls}_k\}$, we firstly use the prompt below to elicit background knowledge from LLMs about each technical term in the abstracts, under the assumption that the most important technical concepts in the full text will also appear in the abstract. The result is a set of explanations $E=\{e_1,e_2,\cdots,e_k\}$.
%to provide sufficient background information for lay summary generation. 
%We use the following prompt to elicit background knowledge from LLMs, resulting in a set of explanations $E=\{e_1,e_2,\cdots,e_k\}$:

\begin{tcolorbox}{
\small
    Generate a thorough background explanation (including definitions, history, and symptoms) of the key biomedical concepts in the following text. 

Text: [\textcolor{blue}{Abstract}]
}
\end{tcolorbox}

Using terms like \textit{thorough} and \textit{definitions} in the prompt encourages responses that are as precise and detailed as possible. We compare two LLMs, i.e., ChatGPT, which is a GPT-3 based chatbot released by OpenAI, and Vicuna-13B which is fine-tuned on LLaMA  \citep{Touvron2023LLaMAOA} using conversational data from ShareGPT\footnote{\url{https://sharegpt.com/}}. 
Vicuna-13B (13 Billion parameters) claims to achieve 90\% of the performance of ChatGPT, but with far less computational cost.
Compared to retrieval-augmented LS methods  \citep{Guo2022}, the use of LLMs in EXpSum presents the advantage of generating superior quality background information, without requiring an external knowledge base or specific training data.

\noindent\textbf{Summarise} Technical source documents in the training set are augmented with the LLM-generated background information prior to training,
to allow the backbone summarisation model to take advantage of this information to generate better quality lay summaries.
Any text generation model may be substituted in this part of the framework, ranging from LLMs e.g., InstructGPT  \citep{Ouyang2022} in
zero-shot or few-shot settings, to PLMs e.g., BART  \citep{Lewis2019}. 
Here, we choose the BART-large version\footnote{\url{huggingface.co/facebook/bart-large-cnn}}, pre-trained on the CNN/DailyMail dataset  \citep{See2017}. 
The model is then fine-tuned on our knowledge-enriched dataset to learn to carry out LS of biomedical text.

\subsection{LLMs for Zero-Shot Lay Summarisation}
\label{sub:ZSLS}
Inspired by the remarkable zero-shot performance of LLMs in text simplification and generation  \citep{lyu2023translating},  we investigate the performance of the same two LLMs (i.e., ChatGPT and Vicuna-13B) in generating complete lay summaries in a zero-shot setting. 
We designed the following prompt which, by enumerating several specific operations, intends to guide the LLMs to generate lay summaries that are as simple as possible:

\begin{tcolorbox}{
\small
    Summarise the following article for a non-expert audience. Please: 1. Replace arcane words with common synonyms. 2. Split long, complex sentences into shorter, simpler sentences. 3. Omit experimental results that are too detailed for lay readers, like confidence intervals and other statistical values. 4. Add explanations for complex terms and abbreviations in the article.
    
Article: [\textcolor{blue}{\textit{Article}}]
}
\end{tcolorbox}
%We specify specific operations instead of abstract principles since we anticipated that this would result in the generation of summaries with a higher degree of layness.

\subsection{LLM for Layness Evaluation}
Traditional readability metrics such as CLI  \citep{Coleman1975}, along with recent PLM-based metrics, only consider individual features as indicators of textual complexity, e.g., the number of letters in a word and the number of words in a sentence. 
To assess summary layness, other perspectives should also be considered, e.g., the extent to which complex terms are avoided.
Motivated by recent work on framing LLMs as automatic evaluators  \citep{Fu2023GPTScoreEA,Luo2023ChatGPTAA}, we propose two LLM-based metrics to assess the layness of generated summaries more effectively than traditional textual complexity measures. 

\noindent\textbf{LLM Rater}
Our first novel metric \textit{LLM Rater} prompts LLMs (ChatGPT and Vicuna) to assess the readability of a summary on a scale of 1-10. 
%in which 10 corresponds to the easiest level of readability. 
The prompt, which is shown below, encourages the LLM to make its judgement based on four layness features, which are derived from the four specific instructions that we used to prompt the LLMs to perform zero-shot LS (see Section \ref{sub:ZSLS}).

\begin{tcolorbox}{
\small
    Score the layness of the following summary from 1 to 10. 10 marks mean the summary is totally easy to understand while 1 mark stands for the summary is the most difficult. Note that layness is the level of ease for laypeople to understand the summary and can be reflected in the following abstracts. 1. To what extent does the summary avoid the use of arcane words? 2. To what extent does the summary avoid the use of technical details that would be difficult for non-expert readers to understand? 3. To what extent does the summary use simple syntactic structures and provide sufficient cohesive cues to allow the text to flow well? 4. To what extent does the summary contain sufficient explanations of any complex terms and abbreviations that are introduced? 
    
    Summary: [\textcolor{blue}{Summary}]
    
    Marks:
}
\end{tcolorbox}

\noindent To facilitate comparison with other metrics, which assign a low score to denote a high level of readability, we take the negative values of LLM-assigned scores plus 10 as the output score of \textit{Rater} metric.

%Rater\textsubscript{GPT} and Rater\textsubscript{Vicuna}  use a lower Since the other metrics that we evaluate assign lower For the convenience of comparison, we take the negative values of LLM-rated scores plus 10 as Rater\textsubscript{GPT} and Rater\textsubscript{Vicuna} in the following passage.

\noindent\textbf{LLM Score} The \textit{LLM Score} is inspired by GPTScore  \citep{Fu2023GPTScoreEA}, which assumes that, given a prompt and context, LLMs will assign a higher probability to a high-quality sequence.
%will be generated by language model, given a prompt, to assess the quality of the attribute described in the prompt. 
The LLM Score corresponds to the cross entropy (CE) loss that is assigned by an LLM to a summary when instructed to summarise an input article with specific simplification operations as in the below template.
\begin{tcolorbox}{
\small
    Generate a lay summary for the following article. Note that layness is the level of ease for
laypeople to understand the summary and
can be reflected in the following abstracts. A lay summary is expected to: 1. Replace arcane words with common synonyms. 2. Split long, complex sentences into shorter, simpler sentences. 3. Omit experimental results that are too detailed for lay readers, like confidence intervals and other statistical values. 4. Add explanations for complex terms and abbreviations in the article.
    
Article: [\textcolor{blue}{\textit{Article}}]

Summary: [\textcolor{blue}{\textit{Summary}}]
}
\end{tcolorbox}
%(see Appendix \ref{appendix: llmscore} for more details). 
%The LLM Score uses the cross entropy (CE) loss from an LLM following the given instruction for lay summarisation with the definition of layness and the input article, as the layness score of a summary.
The underlying assumption is that when instructed to generate text according to a given prefix prompt, LLMs will assign the lowest CE loss to the most likely sequence, which, in our case, corresponds to a summary that exhibits the most lay-like feature. Formally, the \textit{LLM Score} is defined as the follows :
\begin{equation}
    \footnotesize
    LLMSc.(h|d,a,S)=-\sum_{t=1}^mh_tlogp(h_t|h_{<t}, T(d,a,S), \theta)
\end{equation}
where $h$ stands for the evaluated lay summary, $h_t$ is the $t$ th token in $h$, $d$ is the description of the lay summarisation task, $a$ is the target aspect, e.g. layness, and $S$ is the input article. $T$ is a template to combine $d$, $a$, and $S$ into a prefix. 

As ChatGPT does not return with token probabilities, we use only the open-source Vicuna-13B and denote it as LLMS\textsubscript{Vicuna}. 

\section{Experiments}
\textbf{Baseline Approaches.}
We compare the results of ExpSum and zero-shot LS performance of LLMs, with the following baseline approaches: 1) \textbf{BART}  \citep{lewis2020bart}. This method fine-tunes the BART model, using only the technical articles, to generate lay summaries. 2) \textbf{KDR-FT}. The keyword definition retrieval method proposed in   \citet{Guo2022}, which retrieves Wikipedia definitions of the key technical terms in the abstract and prepends them to the article prior to fine-tuning 3) \textbf{DPR-FT}  \citep{Lewis2020RetrievalAugmentedGF}. A dense passage retrieval model, which retrieves snippets related to the input article from the \textit{wiki\_dpr}\footnote{\url{https://huggingface.co/datasets/wiki_dpr}} database. It encodes the article and its five most relevant snippets prior to fine-tuning. For a fair comparison, the three baselines are fine-tuned on the same BART model used by ExpSum.

\textbf{Datasets.}
We use two benchmark datasets for evaluation: 1) \textbf{PLOS}  \citep{Luo2022ReadabilityCB} contains 28,124 abstracts and lay summaries, averaging 287 and 204 words, respectively. 2) \textbf{eLife}  \citep{Goldsack2022} contains 4,828 abstracts and lay summaries, which average 187 words and 386 words, respectively. Lay summaries in eLife typically exhibit a higher degree of linguistic simplicity than those in PLOS   \citep{Goldsack2022}. 

\textbf{Implementation Details.}
To generate background knowledge in ExpSum, we used the ChatGPT API\footnote{\url{https://platform.openai.com/docs/models/gpt-3-5}} and the lmsys/vicuna-13b-delta-v1.1 version Vicuna\footnote{\url{https://huggingface.co/lmsys/vicuna-13b-delta-v1.1}} from Huggingface. The input to fine-tune ExpSum is obtained by truncating the LLM-generated background information to 320 tokens, and the input technical articles to 700 tokens, since the maximum length of texts encoded by the BART backbone is 1024 tokens. 
In terms of decoder parameters, beams are set to 4 and the maximum length of generated sequences is equal to the average length of lay summaries in the dataset. For both the zero-shot LS and LLM-based evaluation metrics, input articles are truncated to 1024 tokens. For the DPR baseline, we used the trained retriever from  \citep{Lewis2020RetrievalAugmentedGF}.

\textbf{Automatic Evaluation Metrics.} To assess the effectiveness of our two LLM-based evaluation metrics, we follow previous approaches by using the Pearson correlation and Spearman correlation  \citep{zar2005spearman} to assess the degree of alignment between the automated metrics and ground truth scores.
Following previous work  \citep{Luo2022ReadabilityCB,Goldsack2022}, generated summaries are evaluated from two different perspectives. i.e., semantics and layness. We use the F1-score of Rouge  \citep{Lin2004} and BertScore  \citep{Zhang2019} to evaluate the semantic similarity between generated and gold-standard summaries. Layness is evaluated by the three textual complexity metrics that correlate best with the ground truth in the benchmark datasets (see Table \ref{tab: Correlation}), i.e., the Coleman–Liau Index (CLI)  \citep{Coleman1975ACR}, our novel Rater\textsubscript{GPT} metric, and the cross entropy of noun phrases (CEoNP) measure. This simplified version of RNPTC  \citep{Luo2022ReadabilityCB}  estimates the layness of a text by averaging the cross entropy loss of predicting every noun phrase. Our implementation CEoNP algorithm can be found in Algorithm \ref{alg:cap}.

\begin{algorithm}
\caption{Given a document $d$ and a PLM $lm$. The FORWARD function takes a corrupt document $d'$ with a masked NP and returns the average cross entropy $ce$ over the tokens within the NP.}\label{alg:cap}
\small
\begin{algorithmic}[1]
\Procedure{CEoNP}{$d, lm$}
\State {NPs $\gets $ Noun phrases list extracted from d}
\State {CE $\gets $ Create empty NP cross entropy list}
\For{$i = \{1,\cdots,|NPs|\}$}
    \State {$T \gets $ Token sequence of $NPi$}
    \State {$d' \gets d$}
    \State {$p \gets $ Create empty token cross entropy list}
    \ForAll{$t \in T$}
        \State {$d'[t] \gets$ [MASK]}
    \EndFor
    \State {$ce \gets$ FORWARD$(lm, d'$)}
    \State {APPEND($CE$, $ce$)}
\EndFor
\State{\textbf{return}  MEAN($CE$)}
\EndProcedure
\end{algorithmic}
\end{algorithm}

%They have shown the highest alignment with the ground truth text complexity levels of PLOS and eLife datasets as shown in Table \ref{tab: Correlation}, compared with other metrics.
%ANPCE achieves 29.74\% Spearman's correlation  \cite{Hauke2011ComparisonOV} and 29.92\% Pearson correlation  \cite{Hauke2011ComparisonOV} on PLOS, 77.60\% Spearman's correlation and 76.07\% Pearson correlation on eLife. Details of this metric and correlation calculation can be found in Appendix.
%The other one is based on GPTScore  \cite{Fu2023GPTScoreEA} which uses the perplexity a language model gives to a sequence conditioned on specific prefixes as the index to note the quality of the sequence. 

\subsection{Results of Automatic Evaluation}
\noindent\textbf{Evaluation Results for Automatic Evaluation Metrics.} 
Table \ref{tab: Correlation} shows the Spearman and Pearson correlations between the scores obtained by different metrics and the ground truth scores for abstracts and summaries in the PLOS and eLife datasets. These ground truth scores are set to 0 for lay summaries and 1 for technical abstracts. 
%Since the PLOS's human-authored lay summaries are generally more complex than eLife's, we can see there are significant discrepancies between the correlation scores for these two datasets.
%The metrics are likely to assign higher scores to the former than the latter.   
The results show that Rater\textsubscript{GPT} achieves the highest correlation with the ground truth scores for both datasets. The performance of Rater\textsubscript{Vicuna} is significantly lower, and LLMS\textsubscript{Vicuna} exhibits the lowest performance of all compared metrics on both corpora. A possible explanation is that since the Vicuna model is trained using only conversational datasets, it is not sufficiently aligned to the task of lay summary evaluation.
%This shows the inefficiency of Vincuna-13B as layness evaluator compared with ChatGPT.
For eLife, the highest correlation is achieved by the traditional CLI metric, which measures readability according to the lengths of words and sentences. The high correlation of this metric is likely to be due to the significant decrease in lexical sophistication and syntactic complexity of lay summaries compared to technical abstracts in eLife. However, the fact that both Rater\textsubscript{GPT} and CEoNP achieve comparable performance to CLI for eLife, and higher performance than CLI for PLOS, highlights the importance of considering a wider range of textual attributes when assessing layness. 
\begin{table}[!t]
    \centering
    \small
    \begin{tabular}{lcccc}
    \hline
        \multirow{2}*{\textbf{Metric}}& \multicolumn{2}{c}{\textbf{PLOS}}&\multicolumn{2}{c}{\textbf{eLife}}\\
         &Spear.&Pear.&Spear.&Pear.\\\hline
         Rater\textsubscript{GPT}&0.272&0.277&0.764&0.775\\
         Rater\textsubscript{Vicuna}&-0.034&-0.031&0.518&0.507\\
         LLMS\textsubscript{Vicuna}&-0.173&-0.172 &0.034&0.045\\
         CEoNP &0.273&0.279&0.749&0.735\\
CLI&0.140&0.136&0.822&0.811\\
    \hline
    \end{tabular}
    \caption{Correlations between evaluation metrics and ground truth textual complexity levels in PLOS and eLife datasets. Spearman's and Pearson correlations are abbreviated as Spear. and Pear.}
    \label{tab: Correlation}
\end{table}

\begin{figure}
    \centering
    \includegraphics[scale=0.32]{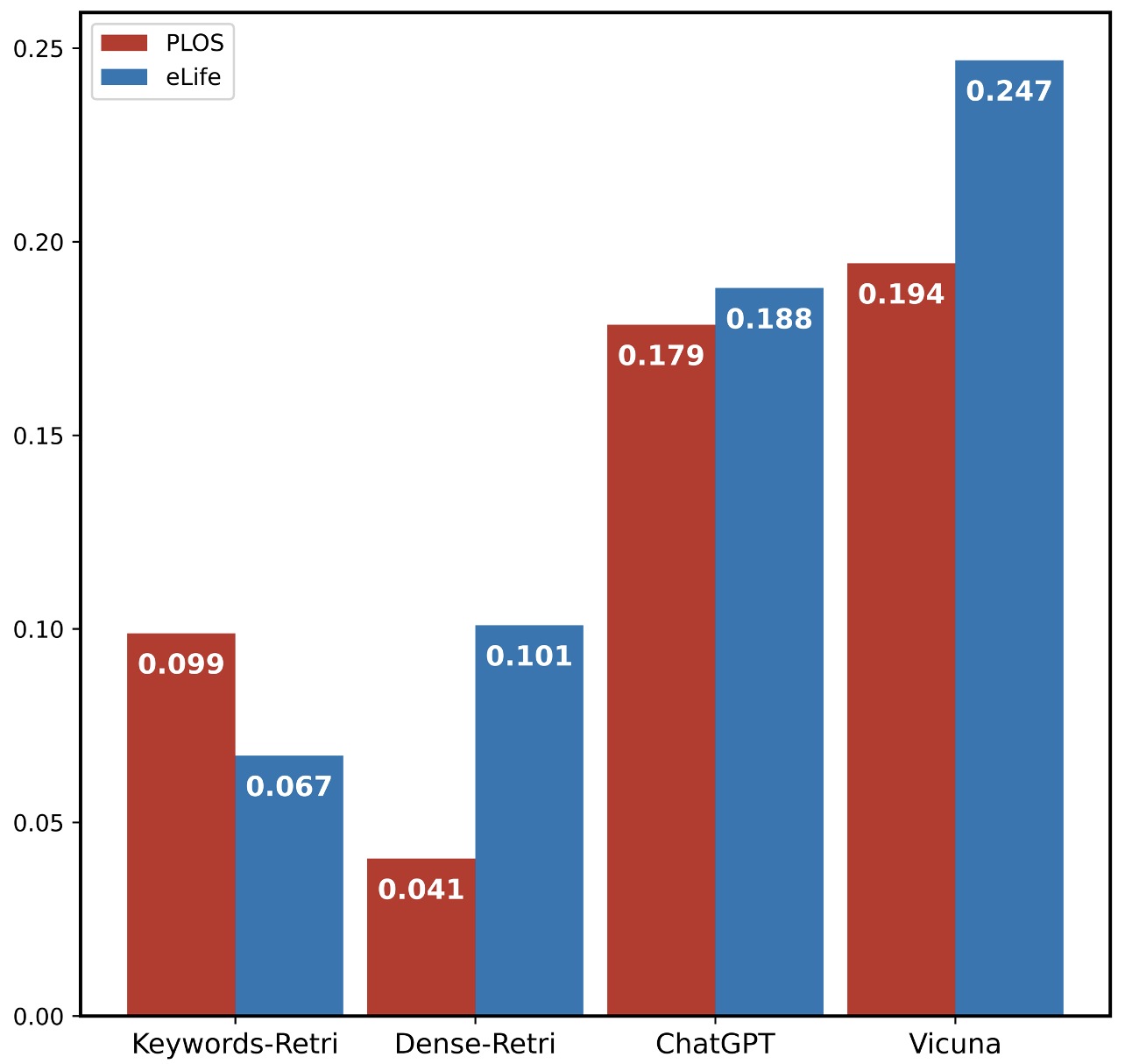}
    \caption{Geometric mean Rouge comparision between different background knowledge acquisition methods.}
    \label{fig: obtained knowledge comparision}
\end{figure}

\noindent\textbf{Evaluation Results for Background Knowledge.} 
Before evaluating the quality of lay summary generation, we firstly evaluate the quality of background knowledge obtained using various different methods. We use the geometric mean of Rouge-1, Rouge-2, and Rouge-L to measure the degree of alignment between the background knowledge obtained by each method and the human-authored lay summaries, as shown in Fig \ref{fig: obtained knowledge comparision}.
The results show that background knowledge generated by LLMs has a much higher degree of semantic similarity to the summary, compared to methods that acquire knowledge from external knowledge bases. Moreover, despite its smaller size, Vicuna's performance surpasses that of ChatGPT on both datasets. This may be due to Vicuna's tendency to produce more verbose term explanations than ChatGPT, according to its fine-tuning on conversational data.

\begin{table*}[!htb]
    \centering
    \small
    \resizebox{\textwidth}{!}{\begin{tabular}{lccccc|ccccc}
    \hline
        \multirow{3}*{\textbf{Methods}}&\multicolumn{5}{c}{\textbf{PLOS}}&\multicolumn{5}{c}{\textbf{eLife}}
        \\
        &\multicolumn{2}{c}{Semantic$\uparrow$}&\multicolumn{3}{c}{Layness$\downarrow$} &\multicolumn{2}{c}{Semantic$\uparrow$}&\multicolumn{3}{c}{Layness$\downarrow$}\\
         & R-1/R-2/R-L&BertS&CLI&CEoNP&Rater\textsubscript{GPT} & R-1/R-2/R-L&BertS&CLI&CEoNP&Rater\textsubscript{GPT}\\\hline
        \textit{SFT} &&&&&&\\
        ExpSum\textsubscript{ChatGPT}&42.35/11.19/38.56&64.49&15.47&6.589&5.992&46.87/11.88/44.08&63.42&11.90&5.265&3.838\\
        ExpSum\textsubscript{Vicuna}&42.84/11.62/38.95 &64.91&15.47 &6.628 &5.959&47.30/12.42/44.49&63.54&11.89&5.250&3.751\\
         KDR-FT &38.95/ 9.52 /35.48&63.01&15.52&6.521&6.269&45.46/10.99/42.86&	62.32&11.66&5.243&3.863\\
        DPR-FT &38.59/ 9.45 /35.10&62.93&15.49&6.444&6.398&45.21/10.77/42.64&62.32&11.91&5.226&3.635\\\hline
        \textit{Zero Shot} &&&&&&\\
        ChatGPT&39.67/10.31/36.20&64.42& 15.40&6.379&4.070&30.77/ 6.56 /28.59&60.86&15.75& 6.512&4.435 \\
        Vicuna& 38.15/ 9.47 /34.91& 63.20&14.16 &6.153 &3.900 &33.50/ 7.05 /31.05 &60.21& 15.11&6.535&4.801 \\\hline
        BART&38.74/ 9.40 /35.20&62.91& 15.26&6.416&6.373&45.36/10.79/	42.86&	62.38&12.00&5.190&3.535\\
        Target &-&-&15.98&6.849&5.506&-&-&12.49&5.845&3.091\\\hline
    \end{tabular}}
    \caption{Lay Summarisation results for both datasets. $SFT$ stands for Supervised FineTune methods.}
    \label{tab:Results}
\end{table*}

\noindent\textbf{Automatic Evaluation of Lay Summarisation.} Table \ref{tab:Results} reports on the  quality of lay summaries generated by different LS methods.
From a semantic perspective (i.e., Rouge and BertScore), ExpSum significantly outperforms other methods,  demonstrating that its use of high quality, LLM-generated background knowledge results in better quality lay summaries, compared to the approaches that retrieve this information from knowledge bases.
%This suggests that the higher quality background information provided by the LLMs is able to bridge the knowledge gap between the technical abstract and the lay summary more effectively than the two retrieval-based approaches, which only slightly enhance the performance on the PLOS dataset and remain close to the baseline performance on the eLife dataset. 
Indeed, the BART method, which does not make use of any background knowledge, performed better than the DPR-FT dense retrieval method, suggesting that background knowledge can actually harm summary quality if it is not of an adequate quality. 
In the zero-shot setting, LLMs are not aware of the content and style of the summaries that they are expected to produce. As such, there is no guarantee that they will generate lay summaries that are similar to human-authored summaries. Nevertheless, on the PLOS dataset, summaries generated by both ChatGPT and Vicuna in the zero-shot setting outperform all other methods apart from ExpSum, in terms of BERTScore. ChatGPT's Rouge also surpasses all supervised fine-tuned models, except for ExpSum. 
The zero-shot performance of LLMs on the eLife dataset is much lower. Because the target lay summaries in this dataset are longer, more abstractive, and more simplified than those in PLOS, zero-shot methods struggle to mimic the required style.  

With regard to layness evaluation (i.e., CLI, CEoNP, and Rater\textsubscript{GPT} scores), the considerably lower scores for eLife indicate that its generated summaries generally exhibit much higher degree of layness. 
For both datasets, there is no discernible difference between the layness levels of summaries generated by any of the supervised fine-tuning models. 
For the PLOS dataset, zero-shot LLM lay summarisation exhibits significantly lower scores than supervised methods, indicating that the more comprehensive summaries generated by LLMs in the zero-shot setting exhibit more lay-like features.
In contrast, applying zero-shot LLM methods to eLife summaries produces summaries that have fewer lay-like features than those generated by supervised methods. This is likely to be because, unlike the zero-shot approaches, the supervised methods can learn the expected level of layness in the target summaries.

Furthermore, it can observed that there are conflicts among the layness metrics. For instance, CLI and CEoNP suggest that target summaries in eLife are more complex than those generated by ExpSum, whereas Rater\textsubscript{GPT} indicates the opposite. As a result, we further investigate the reliability of the automated layness metrics through the human evaluation described below.
%Regarding a comparison of three automated metrics, we can observe Rater\textsubscript{GPT} elicits different implications compared to the other two. For example, the target summaries from eLife achieve higher CLI and CEoNP, but lower Rater\textsubscript{GPT} score, showing the unstable of automated evaluation metrics. 
%Thus, we further conduct the human evaluation to better evaluate the generated lay summaries.
%these are similar to the readability levels of the summaries generated by fine-tuned models for the PLOS dataset, but much higher (i.e., more complex) than the ones generated by fine-tuned models in eLife. This indicates that the lay summaries produced in the zero-shot setting exhibit similar.

\subsection{Human Evaluation}
%While automated metrics provide information about the extent to which generated summaries are semantically similar to target summaries 
Recent studies   \cite{Stiennon2020, Goyal2022} have shown that metrics such as Rouge do not necessarily align well with human assessments regarding the quality and acceptability of summaries generated by LLMs. Accordingly, we have designed a protocol for the human evaluation of lay summaries and recruited assessors to evaluate the quality of generated summaries according to this protocol.

\begin{figure*}[!hbt]
    \centering
    \includegraphics[scale=0.35]{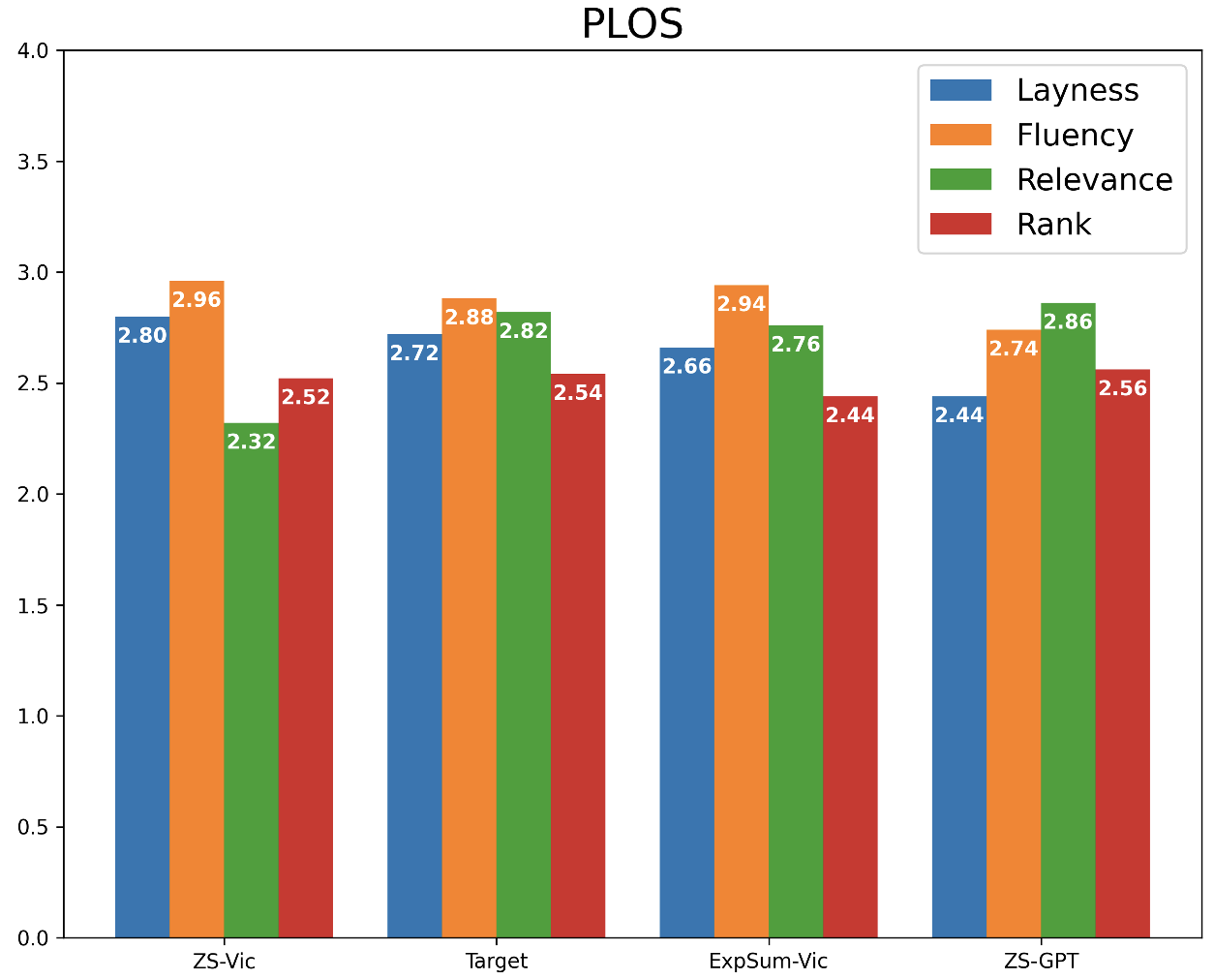}
    \includegraphics[scale=0.35]{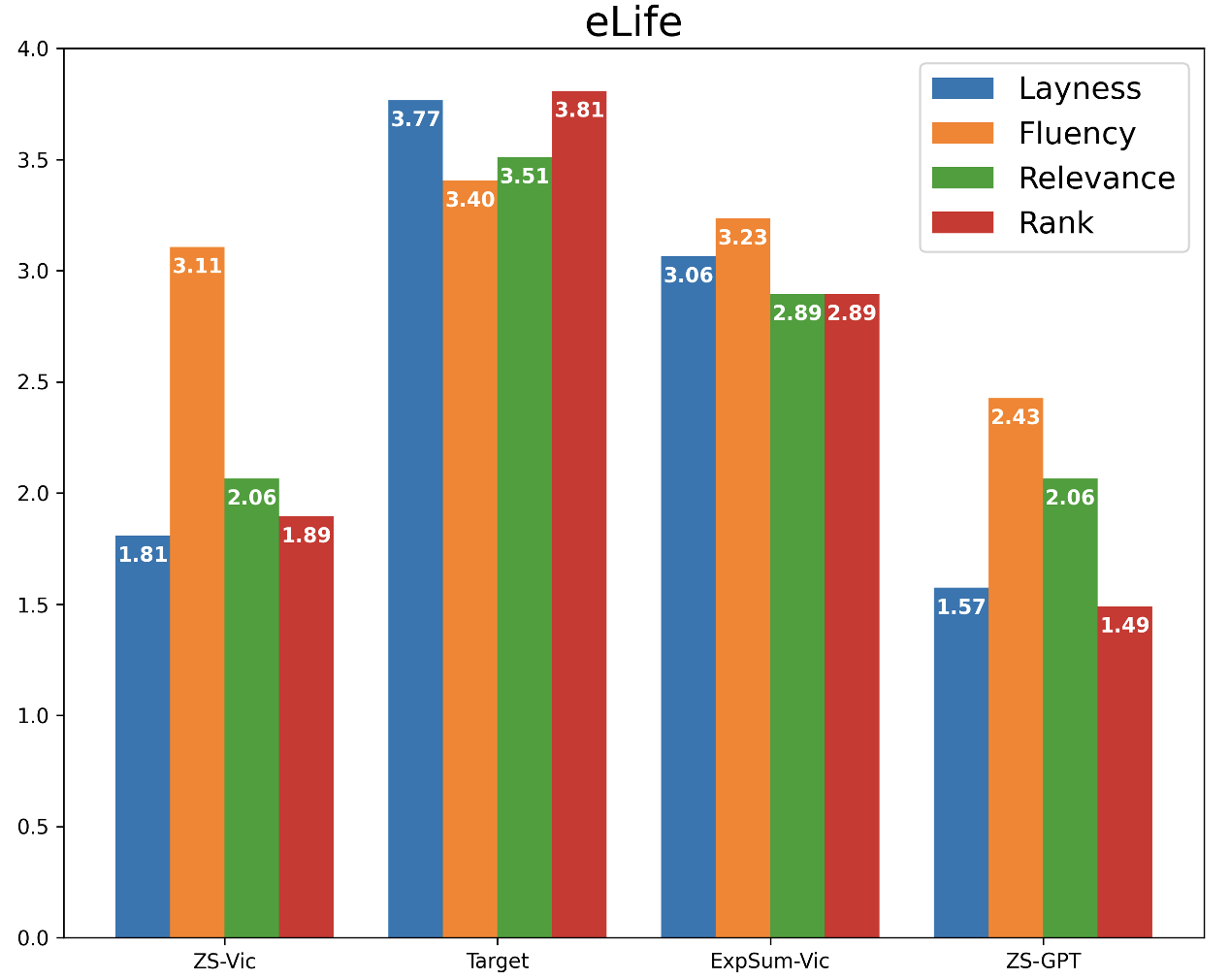}
    \caption{Human Evaluation Results. ZS-Vic and ZS-GPT correspond to zero shot LS with the two respective LLMs. ExpSum-Vic corresponds to ExpSum with background knowledge from Vicuna-13B.}
    \label{fig: Human Evaluation Results}
\end{figure*}

\textbf{Protocol.}
Our protocol encompasses three important aspects for the accurate evaluation of lay summary quality: 1) \textbf{Layness:} Compared to the original article, the summary should decrease the linguistic complexity, omit content that is too technical, and include sufficient background explanation of technical terms. 2) \textbf{Fluency} The summary should present ideas using appropriate lexical and logical connections. 3) \textbf{Relevance:} Although aimed at a non-expert audience, lay summaries should still convey the gist of the source article. Evaluators were asked to mark each aspect on a scale of 1-4, and then rank the summaries generated by different methods based on their overall level of acceptability. 
The detailed assessment protocol is shown as follow:

\textit{Layness}

•	To what extent does the summary use simple words instead of technical jargon? 

•	To what extent does the summary sufficiently omit technical terms (such as statistical significance) that are hard for lay readers to understand?

•	To what extent does the summary use simple syntactic structures (paraphrases) and brief clauses (avoiding subordinate clauses). 

•	How well does the summary explain complex terms and concepts?

\noindent\textit{Fluency}

•	How well does the summary flow? Does it use appropriate grammatical and lexical connections to link parts of the summary?

•	How well do the ideas in the summary flow and present a logical progression?

\noindent\textit{Relevance}

•	How well does the summary convey the key information nuggets of the article? (The abstract of the article is used as the proxy for the gist of the article in the annotation)

\noindent\textit{Overall}:
Please rank the candidate summaries based on their informativeness and your reading experience, then give them an overall mark based on the ranking.
4 marks for the 1st, 3 marks for the 2nd, 2 marks for the 3rd and 1 mark for the 4th.

\textbf{Setup.}
Fifty articles were selected from each of the PLOS and eLife datasets, accompanied by their human-authored lay summary and the lay summaries generated by three different methods, i.e., ExpSum\textsubscript{Vicuna} given its superior performance, along with the zero-shot summaries produced by ChatGPT and Vicuna. To rank the summaries, a score of 4 was assigned to the most acceptable summary, and 1 to the least acceptable summary.  This resulted in a total of 400 summaries being evaluated. 
%Examples of evaluated summaries can be found in Appendix \ref{appendix: lay summary examples}.
%For reasons of efficiency, the relevance aspect is addressed by comparing the lay summary against the original abstract, rather than the full text of the article.  
We recruited four human assessors with either native speakers or near-native competency in English, without a biomedical higher education background. We considered the latter to be important, since domain experts tend not to be able to distinguish the difference in layness between technical and lay summaries  \citep{Luo2022ReadabilityCB}.%Half of the instances were double-scored to allow inter-rater reliability to be assessed. 

\textbf{Analysis of Results.}
The results of the human evaluation are shown in Fig \ref{fig: Human Evaluation Results}. For PLOS, the average  ranking scores are virtually equal for each different type of summary. Furthermore, there are few differences in the scores assigned to the individually assessed aspects across the four types of summaries, with the exception of the relevance score for summaries generated by the ZS-Vic model, which is somewhat lower than for the other types of summaries. This indicates that the human-authored (target) lay summaries do not stand out from the three automatically generated summaries, nor do any of the automated methods perform particularly poorly. In contrast, for eLife, target summaries are much preferred by human assessors over any of the automated methods, with an average ranking score of 3.81. The layness score for these summaries also beats the automatically generated summaries by a significant margin. The benefits of training on target summaries are evidenced by the scores assigned to ExpSum-Vic summaries, most of which are significantly higher than those assigned to two zero-shot methods. While ZS-Vic scores rather well in terms of fluency, both zero-shot methods perform particularly poorly in terms of layness and relevance. The results can once again be explained by the layness discrepancy between the two datasets. While the layness level of the zero-shot LLM methods can equal or even surpass that of the original PLOS lay summaries, this is not the case for eLife summaries. 
%A manual inspection of the evaluated summaries (refer to Appendix \ref{appendix: lay summary examples}) indicates that lay summaries generated by zero-shot LLMs are considerably shorter compared to the eLife target summaries eLife and those generated by ExpSum. Consequently, there is limited space available to include additional explanations to enhance their layness.
%In the zero-shot LS setting, the LLMs do not how simple to make their summaries. However, the layness level of even the supervised method falls short of the layness level of the target summaries in eLife.  This may be because the background explanations generated by Vicuna do not match level of layness of the target eLife summaries. 
%This indicates that the tested LLMs are insufficiently aligned with instructions for conducting lay summarisation.
%In the absence of more advanced LLMs,
Generally, ExpSum-Vic method appears to generate summaries that are acceptable from multiple perspectives, regardless of the level of simplification of the target summaries. 
%result alternative when the desired lay summaries are much more simplified than what current accessible LLMs can produce.
%and it can be improved with In-Context-Learning(ICL) if LLMs with much longer input limits are provided or with better-aligned models like GPT-4. 

Table \ref{tab: Correlations to Human} shows the results of calculating the correlations of scores between the three layness metrics used for automatic evaluation and the human-assigned layness scores. The general pattern is similar to the results in Table \ref{tab: Correlation}, in that Rater\textsubscript{GPT} and CEoNP correlate better with human judgements than CLI in PLOS, while the reverse pattern is observable for eLife. This emphasises that Rater\textsubscript{GPT} and CEoNP are less sensitive to differences in the length of words and sentences. However, the results of the human assessment on layness also demonstrate that Rater\textsubscript{GPT} successfully determines that the target summaries from eLife are more easily comprehensible compared to those generated by ExpSum(see Table \ref{tab:Results}). This finding indicates the significant potential of LLM-rating evaluation methods, as the other two metrics indicate the opposite.

\begin{table}[ht]
\small
    \centering
    \begin{tabular}{ccccc}\hline
        \multirow{2}*{\textbf{Metric}}&\multicolumn{2}{c}{\textbf{PLOS}}&\multicolumn{2}{c}{\textbf{eLife}}\\
        &Pear.&Spear.&Pear.&Spear.\\\hline
        Rater\textsubscript{GPT}&0.152&0.161&0.410&0.380\\
        CEoNP&0.120&0.125&0.452&0.436\\
        CLI&0.110&0.120&0.613&0.642\\\hline
        
    \end{tabular}
    \caption{Correlations of metrics to human evaluated layness.}
    \label{tab: Correlations to Human}
\end{table}

\section{Conclusion}
 In this paper, we have systematically examined how LLMs can be used to support both the generation and evaluation of lay summaries. In terms of generation, we have developed a novel framework, ExpSum, which leverages the capacity of LLMs to generate reliable background information as a means to improve the performance of supervised methods. We have also examined the zero-shot performance of LLMs in generating complete lay summaries. For evaluation, we have proposed two novel LLM-based metrics for evaluating lay summaries. Furthermore, we proposed the first detailed protocol for human evaluation of lay summaries. Our experiments reveal that ExpSum significantly outperforms existing supervised methods, boosted by comprehensive LLM-generated background knowledge. The significant potential of LLMs to carry out zero-shot LS has also been demonstrated since both ChatGPT and Vicuna-13B can compete with supervised methods on certain datasets. Finally, our ChatGPT-based Rater evaluation metric shows strong alignment with human preference. In future work, we will extend our study to consider different contexts and domains and will also investigate the impact of using other LLMs, such as GPT-4, to support LS and evaluation.

\begin{comment}
    \section*{Limitations}
A certain limitation of summarisation is that externally obtained knowledge is not guaranteed to be factual. Contextual information  obtained either from databases or from LLMs does not always stick to facts. If subsequent processing to obtain summaries is carried out, the factuality of generated text can not be guaranteed. However, it is expected future work towards  improving the factuality of LLMs will help to ensure that plain language summarisation contains verified information. 

\section*{Ethics Statement}
Although our proposed ExpSum and zero-shot lay summarisation of LLMs are able to generate plausible lay summaries, the faithfulness or factual consistency of generated summaries cannot be guaranteed. We thus suggest the output of our proposed model should be manually examined by domain experts prior to downstream usage. 
\end{comment}

\bibliographystyle{elsarticle-harv} 
\bibliography{sample}

\begin{thebibliography}{36}
\expandafter\ifx\csname natexlab\endcsname\relax\def\natexlab#1{#1}\fi
\providecommand{\url}[1]{\texttt{#1}}
\providecommand{\href}[2]{#2}
\providecommand{\path}[1]{#1}
\providecommand{\DOIprefix}{doi:}
\providecommand{\ArXivprefix}{arXiv:}
\providecommand{\URLprefix}{URL: }
\providecommand{\Pubmedprefix}{pmid:}
\providecommand{\doi}[1]{\href{http://dx.doi.org/#1}{\path{#1}}}
\providecommand{\Pubmed}[1]{\href{pmid:#1}{\path{#1}}}
\providecommand{\bibinfo}[2]{#2}
\ifx\xfnm\relax \def\xfnm[#1]{\unskip,\space#1}\fi
%Type = Article
\bibitem[{AlKhamissi et~al.(2022)AlKhamissi, Li, Celikyilmaz, Diab and Ghazvininejad}]{AlKhamissi2022ARO}
\bibinfo{author}{AlKhamissi, B.}, \bibinfo{author}{Li, M.}, \bibinfo{author}{Celikyilmaz, A.}, \bibinfo{author}{Diab, M.T.}, \bibinfo{author}{Ghazvininejad, M.}, \bibinfo{year}{2022}.
\newblock \bibinfo{title}{A review on language models as knowledge bases}.
\newblock \bibinfo{journal}{ArXiv} \bibinfo{volume}{abs/2204.06031}.
%Type = Article
\bibitem[{Beltagy et~al.(2020)Beltagy, Peters and Cohan}]{beltagy2020longformer}
\bibinfo{author}{Beltagy, I.}, \bibinfo{author}{Peters, M.E.}, \bibinfo{author}{Cohan, A.}, \bibinfo{year}{2020}.
\newblock \bibinfo{title}{Longformer: The long-document transformer}.
\newblock \bibinfo{journal}{arXiv preprint arXiv:2004.05150} .
%Type = Article
\bibitem[{Brown et~al.(2020)Brown, Mann, Ryder, Subbiah, Kaplan, Dhariwal, Neelakantan, Shyam, Sastry, Askell et~al.}]{brown2020language}
\bibinfo{author}{Brown, T.}, \bibinfo{author}{Mann, B.}, \bibinfo{author}{Ryder, N.}, \bibinfo{author}{Subbiah, M.}, \bibinfo{author}{Kaplan, J.D.}, \bibinfo{author}{Dhariwal, P.}, \bibinfo{author}{Neelakantan, A.}, \bibinfo{author}{Shyam, P.}, \bibinfo{author}{Sastry, G.}, \bibinfo{author}{Askell, A.}, et~al., \bibinfo{year}{2020}.
\newblock \bibinfo{title}{Language models are few-shot learners}.
\newblock \bibinfo{journal}{Advances in neural information processing systems} \bibinfo{volume}{33}, \bibinfo{pages}{1877--1901}.
%Type = Inproceedings
\bibitem[{Chandrasekaran et~al.(2020)Chandrasekaran, Feigenblat, Hovy, Ravichander, Shmueli-Scheuer and de~Waard}]{Chandrasekaran2020OverviewAI}
\bibinfo{author}{Chandrasekaran, M.K.}, \bibinfo{author}{Feigenblat, G.}, \bibinfo{author}{Hovy, E.H.}, \bibinfo{author}{Ravichander, A.}, \bibinfo{author}{Shmueli-Scheuer, M.}, \bibinfo{author}{de~Waard, A.}, \bibinfo{year}{2020}.
\newblock \bibinfo{title}{Overview and insights from the shared tasks at scholarly document processing 2020: Cl-scisumm, laysumm and longsumm}, in: \bibinfo{booktitle}{SDP}.
%Type = Article
\bibitem[{Coleman and Liau(1975a)}]{Coleman1975}
\bibinfo{author}{Coleman, M.}, \bibinfo{author}{Liau, T.L.}, \bibinfo{year}{1975}a.
\newblock \bibinfo{title}{A computer readability formula designed for machine scoring.}
\newblock \bibinfo{journal}{Journal of Applied Psychology} \bibinfo{volume}{60}, \bibinfo{pages}{283--284}.
%Type = Article
\bibitem[{Coleman and Liau(1975b)}]{Coleman1975ACR}
\bibinfo{author}{Coleman, M.}, \bibinfo{author}{Liau, T.L.}, \bibinfo{year}{1975}b.
\newblock \bibinfo{title}{A computer readability formula designed for machine scoring.}
\newblock \bibinfo{journal}{Journal of Applied Psychology} \bibinfo{volume}{60}, \bibinfo{pages}{283--284}.
%Type = Article
\bibitem[{Crossley et~al.(2012)Crossley, Allen and McNamara}]{Crossley2012TextSA}
\bibinfo{author}{Crossley, S.A.}, \bibinfo{author}{Allen, D.}, \bibinfo{author}{McNamara, D.S.}, \bibinfo{year}{2012}.
\newblock \bibinfo{title}{Text simplification and comprehensible input: A case for an intuitive approach}.
\newblock \bibinfo{journal}{Language Teaching Research} \bibinfo{volume}{16}, \bibinfo{pages}{108 -- 89}.
%Type = Article
\bibitem[{Crossley et~al.(2014)Crossley, Yang and McNamara}]{Crossley2014WhatsSS}
\bibinfo{author}{Crossley, S.A.}, \bibinfo{author}{Yang, H.S.}, \bibinfo{author}{McNamara, D.S.}, \bibinfo{year}{2014}.
\newblock \bibinfo{title}{What's so simple about simplified texts? a computational and psycholinguistic investigation of text comprehension and text processing.}
\newblock \bibinfo{journal}{Reading in a foreign language} \bibinfo{volume}{26}, \bibinfo{pages}{92--113}.
%Type = Inproceedings
\bibitem[{Devaraj et~al.(2021)Devaraj, Wallace, Marshall and Li}]{Devaraj2021}
\bibinfo{author}{Devaraj, A.}, \bibinfo{author}{Wallace, B.C.}, \bibinfo{author}{Marshall, I.J.}, \bibinfo{author}{Li, J.J.}, \bibinfo{year}{2021}.
\newblock \bibinfo{title}{Paragraph-level simplification of medical texts}, p. \bibinfo{pages}{4972}.
%Type = Article
\bibitem[{Fu et~al.(2023)Fu, Ng, Jiang and Liu}]{Fu2023GPTScoreEA}
\bibinfo{author}{Fu, J.}, \bibinfo{author}{Ng, S.K.}, \bibinfo{author}{Jiang, Z.}, \bibinfo{author}{Liu, P.}, \bibinfo{year}{2023}.
\newblock \bibinfo{title}{Gptscore: Evaluate as you desire}.
\newblock \bibinfo{journal}{ArXiv} \bibinfo{volume}{abs/2302.04166}.
%Type = Inproceedings
\bibitem[{Goldsack et~al.(2022)Goldsack, Zhang, Lin and Scarton}]{Goldsack2022}
\bibinfo{author}{Goldsack, T.}, \bibinfo{author}{Zhang, Z.}, \bibinfo{author}{Lin, C.}, \bibinfo{author}{Scarton, C.}, \bibinfo{year}{2022}.
\newblock \bibinfo{title}{Making science simple: Corpora for the lay summarisation of scientific literature}.
%Type = Article
\bibitem[{Goyal et~al.(2022)Goyal, Li and Durrett}]{Goyal2022}
\bibinfo{author}{Goyal, T.}, \bibinfo{author}{Li, J.J.}, \bibinfo{author}{Durrett, G.}, \bibinfo{year}{2022}.
\newblock \bibinfo{title}{News summarization and evaluation in the era of gpt-3}.
\newblock \bibinfo{journal}{ArXiv} \bibinfo{volume}{abs/2209.12356}.
%Type = Article
\bibitem[{Guo et~al.(2021)Guo, Ainslie, Uthus, Ontanon, Ni, Sung and Yang}]{Guo2021}
\bibinfo{author}{Guo, M.}, \bibinfo{author}{Ainslie, J.}, \bibinfo{author}{Uthus, D.}, \bibinfo{author}{Ontanon, S.}, \bibinfo{author}{Ni, J.}, \bibinfo{author}{Sung, Y.H.}, \bibinfo{author}{Yang, Y.}, \bibinfo{year}{2021}.
\newblock \bibinfo{title}{Longt5: Efficient text-to-text transformer for long sequences}.
\newblock \bibinfo{journal}{arXiv preprint arXiv:2112.07916} .
%Type = Article
\bibitem[{Guo et~al.(2022)Guo, Qiu, Leroy, Wang and Cohen}]{Guo2022}
\bibinfo{author}{Guo, Y.}, \bibinfo{author}{Qiu, W.}, \bibinfo{author}{Leroy, G.}, \bibinfo{author}{Wang, S.}, \bibinfo{author}{Cohen, T.A.}, \bibinfo{year}{2022}.
\newblock \bibinfo{title}{Cells: A parallel corpus for biomedical lay language generation}.
\newblock \bibinfo{journal}{ArXiv} \bibinfo{volume}{abs/2211.03818}.
%Type = Inproceedings
\bibitem[{Kenton and Toutanova(2019)}]{kenton2019bert}
\bibinfo{author}{Kenton, J.D.M.W.C.}, \bibinfo{author}{Toutanova, L.K.}, \bibinfo{year}{2019}.
\newblock \bibinfo{title}{Bert: Pre-training of deep bidirectional transformers for language understanding}, in: \bibinfo{booktitle}{Proceedings of NAACL-HLT}, pp. \bibinfo{pages}{4171--4186}.
%Type = Article
\bibitem[{Kocmi and Federmann(2023)}]{Kocmi2023}
\bibinfo{author}{Kocmi, T.}, \bibinfo{author}{Federmann, C.}, \bibinfo{year}{2023}.
\newblock \bibinfo{title}{Large language models are state-of-the-art evaluators of translation quality}.
\newblock \bibinfo{journal}{ArXiv} \bibinfo{volume}{abs/2302.14520}.
%Type = Article
\bibitem[{Kojima et~al.(2022)Kojima, Gu, Reid, Matsuo and Iwasawa}]{Kojima2022}
\bibinfo{author}{Kojima, T.}, \bibinfo{author}{Gu, S.S.}, \bibinfo{author}{Reid, M.}, \bibinfo{author}{Matsuo, Y.}, \bibinfo{author}{Iwasawa, Y.}, \bibinfo{year}{2022}.
\newblock \bibinfo{title}{Large language models are zero-shot reasoners}.
\newblock \bibinfo{journal}{ArXiv} \bibinfo{volume}{abs/2205.11916}.
%Type = Article
\bibitem[{Lewis et~al.(2019)Lewis, Liu, Goyal, Ghazvininejad, Mohamed, Levy, Stoyanov and Zettlemoyer}]{Lewis2019}
\bibinfo{author}{Lewis, M.}, \bibinfo{author}{Liu, Y.}, \bibinfo{author}{Goyal, N.}, \bibinfo{author}{Ghazvininejad, M.}, \bibinfo{author}{Mohamed, A.}, \bibinfo{author}{Levy, O.}, \bibinfo{author}{Stoyanov, V.}, \bibinfo{author}{Zettlemoyer, L.}, \bibinfo{year}{2019}.
\newblock \bibinfo{title}{Bart: Denoising sequence-to-sequence pre-training for natural language generation, translation, and comprehension}.
\newblock \bibinfo{journal}{arXiv preprint arXiv:1910.13461} .
%Type = Inproceedings
\bibitem[{Lewis et~al.(2020a)Lewis, Liu, Goyal, Ghazvininejad, Mohamed, Levy, Stoyanov and Zettlemoyer}]{lewis2020bart}
\bibinfo{author}{Lewis, M.}, \bibinfo{author}{Liu, Y.}, \bibinfo{author}{Goyal, N.}, \bibinfo{author}{Ghazvininejad, M.}, \bibinfo{author}{Mohamed, A.}, \bibinfo{author}{Levy, O.}, \bibinfo{author}{Stoyanov, V.}, \bibinfo{author}{Zettlemoyer, L.}, \bibinfo{year}{2020}a.
\newblock \bibinfo{title}{Bart: Denoising sequence-to-sequence pre-training for natural language generation, translation, and comprehension}, in: \bibinfo{booktitle}{Proceedings of the 58th Annual Meeting of the Association for Computational Linguistics}, pp. \bibinfo{pages}{7871--7880}.
%Type = Article
\bibitem[{Lewis et~al.(2020b)Lewis, Perez, Piktus, Petroni, Karpukhin, Goyal, Kuttler, Lewis, tau Yih, Rockt{\"a}schel, Riedel and Kiela}]{Lewis2020RetrievalAugmentedGF}
\bibinfo{author}{Lewis, P.}, \bibinfo{author}{Perez, E.}, \bibinfo{author}{Piktus, A.}, \bibinfo{author}{Petroni, F.}, \bibinfo{author}{Karpukhin, V.}, \bibinfo{author}{Goyal, N.}, \bibinfo{author}{Kuttler, H.}, \bibinfo{author}{Lewis, M.}, \bibinfo{author}{tau Yih, W.}, \bibinfo{author}{Rockt{\"a}schel, T.}, \bibinfo{author}{Riedel, S.}, \bibinfo{author}{Kiela, D.}, \bibinfo{year}{2020}b.
\newblock \bibinfo{title}{Retrieval-augmented generation for knowledge-intensive nlp tasks}.
\newblock \bibinfo{journal}{ArXiv} \bibinfo{volume}{abs/2005.11401}.
%Type = Inproceedings
\bibitem[{Lin(2004)}]{Lin2004}
\bibinfo{author}{Lin, C.Y.}, \bibinfo{year}{2004}.
\newblock \bibinfo{title}{Rouge: A package for automatic evaluation of summaries}, pp. \bibinfo{pages}{74--81}.
%Type = Inproceedings
\bibitem[{Liu et~al.(2021)Liu, Liu, Lu, Welleck, West, Bras, Choi and Hajishirzi}]{Liu2021GeneratedKP}
\bibinfo{author}{Liu, J.}, \bibinfo{author}{Liu, A.}, \bibinfo{author}{Lu, X.}, \bibinfo{author}{Welleck, S.}, \bibinfo{author}{West, P.}, \bibinfo{author}{Bras, R.L.}, \bibinfo{author}{Choi, Y.}, \bibinfo{author}{Hajishirzi, H.}, \bibinfo{year}{2021}.
\newblock \bibinfo{title}{Generated knowledge prompting for commonsense reasoning}, in: \bibinfo{booktitle}{Annual Meeting of the Association for Computational Linguistics}.
%Type = Inproceedings
\bibitem[{Luo et~al.(2022)Luo, Xie and Ananiadou}]{Luo2022ReadabilityCB}
\bibinfo{author}{Luo, Z.}, \bibinfo{author}{Xie, Q.}, \bibinfo{author}{Ananiadou, S.}, \bibinfo{year}{2022}.
\newblock \bibinfo{title}{Readability controllable biomedical document summarization}, in: \bibinfo{booktitle}{Conference on Empirical Methods in Natural Language Processing}.
%Type = Inproceedings
\bibitem[{Luo et~al.(2023)Luo, Xie and Ananiadou}]{Luo2023ChatGPTAA}
\bibinfo{author}{Luo, Z.}, \bibinfo{author}{Xie, Q.}, \bibinfo{author}{Ananiadou, S.}, \bibinfo{year}{2023}.
\newblock \bibinfo{title}{Chatgpt as a factual inconsistency evaluator for text summarization}.
%Type = Article
\bibitem[{Lyu et~al.(2023)Lyu, Tan, Zapadka, Ponnatapuram, Niu, Wang and Whitlow}]{lyu2023translating}
\bibinfo{author}{Lyu, Q.}, \bibinfo{author}{Tan, J.}, \bibinfo{author}{Zapadka, M.E.}, \bibinfo{author}{Ponnatapuram, J.}, \bibinfo{author}{Niu, C.}, \bibinfo{author}{Wang, G.}, \bibinfo{author}{Whitlow, C.T.}, \bibinfo{year}{2023}.
\newblock \bibinfo{title}{Translating radiology reports into plain language using chatgpt and gpt-4 with prompt learning: Promising results, limitations, and potential}.
\newblock \bibinfo{journal}{arXiv preprint arXiv:2303.09038} .
%Type = Article
\bibitem[{Ouyang et~al.(2022)Ouyang, Wu, Jiang, Almeida, Wainwright, Mishkin, Zhang, Agarwal, Slama, Ray, Schulman, Hilton, Kelton, Miller, Simens, Askell, Welinder, Christiano, Leike and Lowe}]{Ouyang2022}
\bibinfo{author}{Ouyang, L.}, \bibinfo{author}{Wu, J.}, \bibinfo{author}{Jiang, X.}, \bibinfo{author}{Almeida, D.}, \bibinfo{author}{Wainwright, C.L.}, \bibinfo{author}{Mishkin, P.}, \bibinfo{author}{Zhang, C.}, \bibinfo{author}{Agarwal, S.}, \bibinfo{author}{Slama, K.}, \bibinfo{author}{Ray, A.}, \bibinfo{author}{Schulman, J.}, \bibinfo{author}{Hilton, J.}, \bibinfo{author}{Kelton, F.}, \bibinfo{author}{Miller, L.E.}, \bibinfo{author}{Simens, M.}, \bibinfo{author}{Askell, A.}, \bibinfo{author}{Welinder, P.}, \bibinfo{author}{Christiano, P.F.}, \bibinfo{author}{Leike, J.}, \bibinfo{author}{Lowe, R.J.}, \bibinfo{year}{2022}.
\newblock \bibinfo{title}{Training language models to follow instructions with human feedback}.
\newblock \bibinfo{journal}{ArXiv} \bibinfo{volume}{abs/2203.02155}.
%Type = Article
\bibitem[{Plavén-Sigray et~al.(2017)Plavén-Sigray, Matheson, Schiffler and Thompson}]{10.7554/eLife.27725}
\bibinfo{author}{Plavén-Sigray, P.}, \bibinfo{author}{Matheson, G.J.}, \bibinfo{author}{Schiffler, B.C.}, \bibinfo{author}{Thompson, W.H.}, \bibinfo{year}{2017}.
\newblock \bibinfo{title}{Research: The readability of scientific texts is decreasing over time}.
\newblock \bibinfo{journal}{eLife} \bibinfo{volume}{6}, \bibinfo{pages}{e27725}.
\newblock \URLprefix \url{https://doi.org/10.7554/eLife.27725}, \DOIprefix\doi{10.7554/eLife.27725}.
%Type = Article
\bibitem[{See et~al.(2017)See, Liu and Manning}]{See2017}
\bibinfo{author}{See, A.}, \bibinfo{author}{Liu, P.J.}, \bibinfo{author}{Manning, C.D.}, \bibinfo{year}{2017}.
\newblock \bibinfo{title}{Get to the point: Summarization with pointer-generator networks}.
\newblock \bibinfo{journal}{arXiv preprint arXiv:1704.04368} .
%Type = Inproceedings
\bibitem[{Shaib et~al.(2023)Shaib, Li, Joseph, Marshall, Li and Wallace}]{Shaib2023SummarizingSA}
\bibinfo{author}{Shaib, C.}, \bibinfo{author}{Li, M.}, \bibinfo{author}{Joseph, S.}, \bibinfo{author}{Marshall, I.J.}, \bibinfo{author}{Li, J.J.}, \bibinfo{author}{Wallace, B.}, \bibinfo{year}{2023}.
\newblock \bibinfo{title}{Summarizing, simplifying, and synthesizing medical evidence using gpt-3 (with varying success)}.
%Type = Article
\bibitem[{Shifman et~al.(2008)Shifman, Johannesson, Bronstein, Chen, Collier, Craddock, Kendler, Li, O’Donovan, O’Neill, Owen, Walsh, Weinberger, Sun, Flint and Darvasi}]{Shifman2008GenomeWideAI}
\bibinfo{author}{Shifman, S.}, \bibinfo{author}{Johannesson, M.}, \bibinfo{author}{Bronstein, M.}, \bibinfo{author}{Chen, S.X.}, \bibinfo{author}{Collier, D.A.}, \bibinfo{author}{Craddock, N.}, \bibinfo{author}{Kendler, K.S.}, \bibinfo{author}{Li, T.}, \bibinfo{author}{O’Donovan, M.C.}, \bibinfo{author}{O’Neill, F.A.}, \bibinfo{author}{Owen, M.J.}, \bibinfo{author}{Walsh, D.}, \bibinfo{author}{Weinberger, D.R.}, \bibinfo{author}{Sun, C.}, \bibinfo{author}{Flint, J.}, \bibinfo{author}{Darvasi, A.}, \bibinfo{year}{2008}.
\newblock \bibinfo{title}{Genome-wide association identifies a common variant in the reelin gene that increases the risk of schizophrenia only in women}.
\newblock \bibinfo{journal}{PLoS Genetics} \bibinfo{volume}{4}.
%Type = Article
\bibitem[{Stiennon et~al.(2020)Stiennon, Ouyang, Wu, Ziegler, Lowe, Voss, Radford, Amodei and Christiano}]{Stiennon2020}
\bibinfo{author}{Stiennon, N.}, \bibinfo{author}{Ouyang, L.}, \bibinfo{author}{Wu, J.}, \bibinfo{author}{Ziegler, D.M.}, \bibinfo{author}{Lowe, R.J.}, \bibinfo{author}{Voss, C.}, \bibinfo{author}{Radford, A.}, \bibinfo{author}{Amodei, D.}, \bibinfo{author}{Christiano, P.}, \bibinfo{year}{2020}.
\newblock \bibinfo{title}{Learning to summarize from human feedback}.
\newblock \bibinfo{journal}{ArXiv} \bibinfo{volume}{abs/2009.01325}.
%Type = Article
\bibitem[{Touvron et~al.(2023)Touvron, Lavril, Izacard, Martinet, Lachaux, Lacroix, Rozi{\`e}re, Goyal, Hambro, Azhar, Rodriguez, Joulin, Grave and Lample}]{Touvron2023LLaMAOA}
\bibinfo{author}{Touvron, H.}, \bibinfo{author}{Lavril, T.}, \bibinfo{author}{Izacard, G.}, \bibinfo{author}{Martinet, X.}, \bibinfo{author}{Lachaux, M.A.}, \bibinfo{author}{Lacroix, T.}, \bibinfo{author}{Rozi{\`e}re, B.}, \bibinfo{author}{Goyal, N.}, \bibinfo{author}{Hambro, E.}, \bibinfo{author}{Azhar, F.}, \bibinfo{author}{Rodriguez, A.}, \bibinfo{author}{Joulin, A.}, \bibinfo{author}{Grave, E.}, \bibinfo{author}{Lample, G.}, \bibinfo{year}{2023}.
\newblock \bibinfo{title}{Llama: Open and efficient foundation language models}.
\newblock \bibinfo{journal}{ArXiv} \bibinfo{volume}{abs/2302.13971}.
%Type = Article
\bibitem[{Yu et~al.(2022a)Yu, Iter, Wang, Xu, Ju, Sanyal, Zhu, Zeng and Jiang}]{yu2022generate}
\bibinfo{author}{Yu, W.}, \bibinfo{author}{Iter, D.}, \bibinfo{author}{Wang, S.}, \bibinfo{author}{Xu, Y.}, \bibinfo{author}{Ju, M.}, \bibinfo{author}{Sanyal, S.}, \bibinfo{author}{Zhu, C.}, \bibinfo{author}{Zeng, M.}, \bibinfo{author}{Jiang, M.}, \bibinfo{year}{2022}a.
\newblock \bibinfo{title}{Generate rather than retrieve: Large language models are strong context generators}.
\newblock \bibinfo{journal}{arXiv preprint arXiv:2209.10063} .
%Type = Article
\bibitem[{Yu et~al.(2022b)Yu, Iter, Wang, Xu, Ju, Sanyal, Zhu, Zeng and Jiang}]{Yu2022GenerateRT}
\bibinfo{author}{Yu, W.}, \bibinfo{author}{Iter, D.}, \bibinfo{author}{Wang, S.}, \bibinfo{author}{Xu, Y.}, \bibinfo{author}{Ju, M.}, \bibinfo{author}{Sanyal, S.}, \bibinfo{author}{Zhu, C.}, \bibinfo{author}{Zeng, M.}, \bibinfo{author}{Jiang, M.}, \bibinfo{year}{2022}b.
\newblock \bibinfo{title}{Generate rather than retrieve: Large language models are strong context generators}.
\newblock \bibinfo{journal}{ArXiv} \bibinfo{volume}{abs/2209.10063}.
%Type = Article
\bibitem[{Zar(2005)}]{zar2005spearman}
\bibinfo{author}{Zar, J.H.}, \bibinfo{year}{2005}.
\newblock \bibinfo{title}{Spearman rank correlation}.
\newblock \bibinfo{journal}{Encyclopedia of biostatistics} \bibinfo{volume}{7}.
%Type = Article
\bibitem[{Zhang et~al.(2019)Zhang, Merck, Tsai, Manning and Langlotz}]{Zhang2019}
\bibinfo{author}{Zhang, Y.}, \bibinfo{author}{Merck, D.}, \bibinfo{author}{Tsai, E.B.}, \bibinfo{author}{Manning, C.D.}, \bibinfo{author}{Langlotz, C.P.}, \bibinfo{year}{2019}.
\newblock \bibinfo{title}{Optimizing the factual correctness of a summary: A study of summarizing radiology reports}.
\newblock \bibinfo{journal}{arXiv preprint arXiv:1911.02541} \bibinfo{note}{Factual reward : parser input sentence, pick wanted variable to form a vector, then use RL to optimize it}.

\end{thebibliography}


%\begin{thebibliography}{00}

%% \bibliography{sample}

%\end{thebibliography}
\end{document}